# A Study of an Modeling Method of T-S fuzzy System Based on Moving Fuzzy Reasoning and Its Application

**Son-Il Kwak**

College of Computer Science,　Kim Il Sung University,

Pyongyang, DPR of Korea

**Gang Choe**

College of Computer Science,　Kim Il Sung University,

Pyongyang, DPR of Korea

**In-Song Kim**

Electronic Library,　Kim Il Sung University,

Pyongyang, DPR of Korea

**Gyong-Ho Jo**

College of Computer Science,　Kim Il Sung University,

Pyongyang, DPR of Korea

**Chol-Jun Hwang**

Electronic Library,　Kim Il Sung University,

Pyongyang, DPR of Korea

**Abstract-** To improve the effectiveness of the fuzzy identification, a structure identification method based on moving rate is proposed for T-S fuzzy model. The proposed method is called "T-S modeling (or T-S fuzzy identification method) based on moving rate". First, to improve the shortcomings of existing fuzzy reasoning methods based on matching degree, the moving rates for *s*-type, *z*-type and *trapezoidal* membership functions of T-S fuzzy model were defined. Then, the differences between proposed moving rate and existing matching degree were explained. Next, the identification method based on moving rate was proposed for T-S model. Finally, the proposed identification method was applied to the fuzzy modeling for the precipitation forecast and security situation prediction. Test results show that the proposed method significantly improves the effectiveness of fuzzy identification.

**1. Introduction**

It has been proved that fuzzy systems are useful to simulate a nonlinear system and control. There are mainly two kinds of rule-based fuzzy models: *Mamdani* fuzzy model and *Takagi–Sugeno* (T-S) fuzzy model. The main difference between them is that the consequence parts of *Mamdani* fuzzy model are fuzzy sets while those of the T-S fuzzy model are linear functions of input variables. Compared with *Mamdani* fuzzy model, T-S fuzzy model can approximate complex nonlinear systems with fewer rules and higher modeling accuracy. Therefore, it becomes an active research area. Among the research issues in T-S fuzzy modeling, the identification is the most important and critical one [1]. For the reason that it employs linear model in the consequence part, conventional linear system theory can be applied to the system analysis and synthesis accordingly. Hence, the T–S fuzzy models are becoming powerful engineering tools for modeling and control systems [2].

In identification of T–S fuzzy models, the structure determination and parameter identification are often concerned, i.e., determination of the premise and consequence variables and identification of the premise and consequence parameters [3]. Structure identification of fuzzy model is to determine the number of rules and parameter estimation. The task of structure identification and parameter estimation of T-S fuzzy model is to partition the input space into a number of fuzzy subspaces and obtain parameters.

There are many methods of structure and parameter identification in fuzzy modeling. *Chuen-Tsai Sun* proposed a hill-climbing method based on k-d trees to solve the structure identification problem [4]. Other identification methods have also been proposed. For example, the one which combined the clustering-based fuzzy modeling method [5, 6, 7, 8, 9], the approaches of neural-fuzzy identification [10, 11, 12, 13, 14, 15], GA-based fuzzy identification approaches [16, 17, 2] and so on [18, 19, 20, 21]. However, these identification methods are mainly based on the matching degree-type fuzzy reasoning method.

The fuzzy reasoning is an important tool that should process the input information and should determine a new decision, and fuzzy rules are key factors that guarantee the stability and quality of the system. In addition, the membership function of fuzzy rules is the fundamental element influencing the result of fuzzy reasoning. *Sugeno* and *Kang* discussed the problems of structure identification of a fuzzy model in 1988 [22]. In this method, the main task is to obtain the structure and parameters of fuzzy models. So the method is widely applied in fuzzy modeling. However, there is no change in the calculation results for the grade of the membership function when the membership function of fuzzy set is changed.

There are disadvantages of matching degree-type fuzzy reasoning method that if the matching degrees are the same for the different inputs, sub reasoning results will also be the same. When it comes to the input is fuzzy singleton (crisp), the sub reasoning results will be the same if the values of membership function for the input are the same. In case that the input is a fuzzy set, if the value of matching degree for the premise is the same, it will still be the same sub reasoning results despite of different membership functions. Therefore, the validity of this method is rather low. The reason of misleading reasoning result is that the calculation of membership function is based on considering the geometrical center as the center of fuzzy set regardless whether the membership types is symmetric or asymmetric. Therefore, it is necessary to establish new fuzzy reasoning method to overcome the limitation of the matching degree-type fuzzy reasoning.

With this aspect taken into consideration, in this paper, we define a concept of moving rate of *s*-type, *z*-type and *trapezoidal* membership function, and compare the efficiency between existing methods and the proposed method. In addition, the effectiveness of the proposed method is proved by applying the experiment for precipitation forecast and security situation.

**2. The definition of moving rate and comparison with existing method**

*2.1. T–S fuzzy model*

T-S fuzzy model is expressed with Eq. (1) [2]

$$L^i : if \quad x_1 \quad is \quad A_1^i, \quad x_2 \quad is \quad A_2^i, \ldots, \quad x_m \quad is \quad A_m^i \quad then \quad y^i = a_0^i + \sum_{j=1}^{m} a_j^i x_j \tag{1}$$

where, $L^i (i=1,2,\cdots,n)$ is *i*-th fuzzy rule, $x_j (j=1,2,\cdots,m)$ is input variable and $y^i$ is output variable obtained by $L^i$.

Given the input $x_1^0, x_2^0, \ldots, x_j^0, \ldots, x_m^0$, the final output of the fuzzy model is inferred by a weighted average defuzzification as Eq. (2)

$$\hat{y} = \frac{\sum_{i=1}^{n} w^i y^i}{\sum_{i=1}^{n} w^i} \tag{2}$$

where the weight $w^i$ implies the overall truth value of the premise of the *j-th* implication for the input, and is calculated as Eq. (3)

$$w^i = \prod_{j=1}^{m} \mu_{A_i}(x_j) \tag{3}$$

where $\mu_{A_i}(x_j)$ is the grade of the membership function (MF) of $x_j$ in fuzzy set $A_i$ and is characterized by a Gaussian function as Eq. (4).

$$\mu_{A_i}(x_j) = \exp(-\frac{(x_j - c_i^j)^2}{b_i^j}) \quad . \tag{4}$$

When desired input–output data pairs are given, which are derived or acquired from an unknown function or a system, the problem of fuzzy modeling is how to build a fuzzy model that can properly approximate the input–output relationship. Thus, the task is to design a fuzzy model that consists of finding an optimized structure (number of rules and inputs) and optimizing MF parameters so that the error between the fuzzy model output and the desired output can be minimized.

## 2.2. Moving rate

The moving rates for *s*-type, *z*-type and *trapezoidal* membership functions are defined. Figure 1, 2 and 3 show three types of membership function.

Given a fuzzy rule of Eq. (1), the moving rate $d_{ij}(x_{j0})$ for *z*-type membership function shown in Figure 1 is defined as Eq. (5).

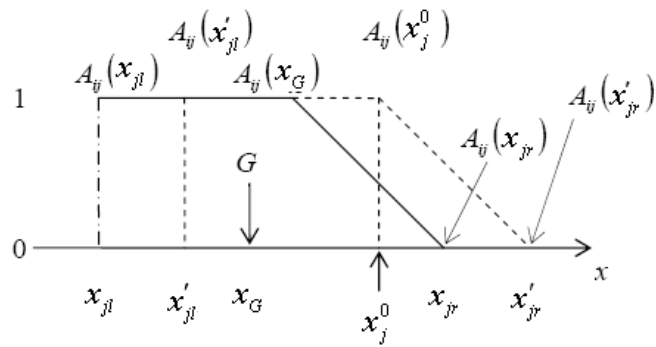

Figure 1. *z*-type membership function and moving rate

$$d_{ij}\left(x_j^0\right)=\begin{cases} \dfrac{x_j^0 - x_G}{x_{jr} - x_{jl}}, & \text{if } x_{jr} \geq x_j^0 \geq x_G \\ \dfrac{x_G - x_j^0}{x_{jr} - x_{jl}}, & \text{if } x_{jl} \leq x_j^0 \leq x_G \\ 0, & \text{if } x_j^0 < x_{jl},\ x_j^0 > x_{jr} \end{cases} \quad (5)$$

In Eq. (5), $x_j^0$ is the input and $x_G$ is the coordinate of the weight center of *z*-type membership function. In addition, $x_{jl}$ is the left endpoint and $x_{jr}$ is right endpoint.

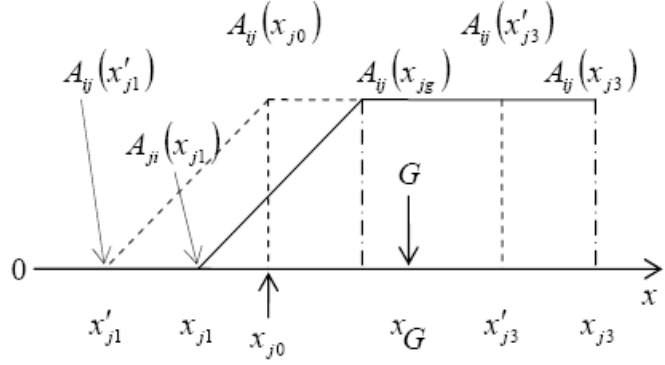

Figure 2. *s*-type membership function and moving rate

Given a fuzzy rule of Eq. (1), the moving rate for *s*-type membership function shown in Figure 2 is defined as Eq. (6).

$$d_{ij}(x_j^0) = \begin{cases} \dfrac{x_G - x_j^0}{x_{jr} - x_{jl}}, & \text{if } x_G \geq x_j^0 \geq x_{jl} \\ \dfrac{x_j^0 - x_G}{x_{jr} - x_{jl}}, & \text{if } x_{jr} \geq x_j^0 \geq x_G \\ 0, & \text{if } x_j^0 > x_{jr},\ x_j^0 < x_{jl} \end{cases} \quad (6)$$

In Eq. (6), $x_j^0$ is the input and $x_G$ is the coordinate of the weight center of *s*-type membership function. In addition, $x_{jl}$ is the left endpoint and $x_{jr}$ is right endpoint.

Given a fuzzy rule of Eq. (1), the moving rate for *trapezoidal* membership shown in Figure 3 is defined as Eq. (7).

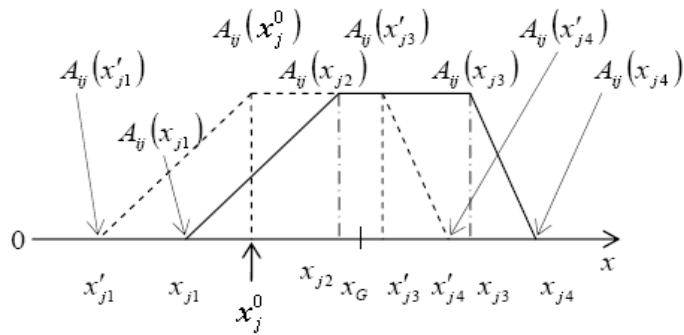

Figure 3. *Trapezoidal* membership function and moving rate

$$d_{ji}(x_j^0) = \begin{cases} \dfrac{x_{j2} - x_j^0}{x_{j4} - x_{j1}}, & \text{if } x_{j2} \geq x_j^0 \geq x_{j1} \\ \dfrac{x_G - x_j^0}{x_{j4} - x_{j1}}, & \text{if } x_{j2} \leq x_j^0 \leq x_G \\ \dfrac{x_j^0 - x_G}{x_{j4} - x_{j1}}, & \text{if } x_G \leq x_j^0 \leq x_{j3} \\ \dfrac{x_j^0 - x_{j3}}{x_{j4} - x_{j1}}, & \text{if } x_{j3} \leq x_j^0 \leq x_{j4} \\ 0, & \text{if } x_{j0} < x_{j1},\ x_j^0 > x_{j4} \end{cases} \quad (7)$$

In Eq. (7), $x_{j0}$ is the input and $x_G$ is the coordinate of the weight center of *trapezoidal* membership function. In addition, $x_{j1}$ and $x_{j4}$ are the coordinates of which the grade are 0, $x_{j2}$ and $x_{j3}$ are the coordinates of which the grade are 1. Eq. (7) shows the moving rate when the *trapezoidal* membership function $A_{ji}(x_{j1})$, $A_{ji}(x_{j2})$, $A_{ji}(x_{j3})$, $A_{ji}(x_{j4})$ are moved to new trapezoidal membership function $A_{ji}(x'_{j1})$, $A_{ji}(x_j^0)$, $A_{ji}(x'_{j3})$, $A_{ji}(x'_{j4})$.

### *2.3. Comparison with existing method*

Moving rate, which were defined in Eq. (5), (6) and (7), considered the center of fuzzy set as the weight center, not geometrical center, so it exactly reflects the grade of the membership function of input and the fuzzy set. But, in [22, 23], the grade of the membership function was defined as height of crossing point between input and membership function, so it doesn't exactly express the grade of the membership function in case of membership function change. For example, consider the situation in Figure 4.

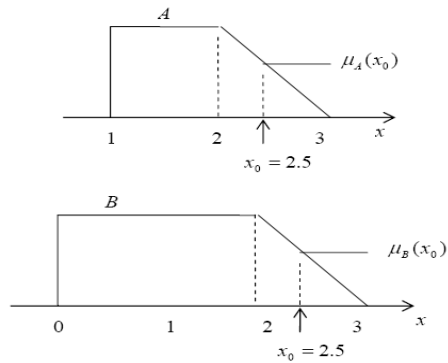

Figure 4. Relationship between grade of the membership function and moving rate

In Figure 4, the membership functions *A* and *B* are different ones. Therefore, generally speaking, the grade of the membership function may be different. But in existing method,

$$\mu_A(x_0) = \mu_A(2.5) = 0.5, \quad \mu_B(x_0) = \mu_B(2.5) = 0.5.$$

The corresponding grade of the membership function may frequently be the same for different membership functions.

On the other hand, in proposed method, weight center of membership function *A* is 1.777, and *B* is 1.266 6. The results are different. In Eq. (5), moving rate of two membership functions can be calculated as follows.

$$d_A(x_0) = 0.3615, \quad d_B(x_0) = 0.4111.$$

That is, by proposed method, different moving rates are calculated for different membership function. The proposed method is a reasoning method that different reasoning results from different membership function and input information are obtained.

## 3. T-S Fuzzy modeling based on moving rate

### *3.1. Fuzzy reasoning method and identification method based on moving rate*

As you can see in Eq. (1), the object to identify is a multi-input single-output (MISO) system.

For input $x_1^0, x_2^0, ..., x_j^0, ..., x_m^0$, the output $y^0$ by *n* rules based on moving rate can be obtained by Eq. (8)

$$y^0 = \frac{\sum_{i=1}^{n} d^i y^i}{\sum_{i=1}^{n} d^i} \tag{8}$$

where, $d_i$ is weight of *i*-th rule defined as Eq. (9)

$$d^i = \prod_{j=1}^{m} d_{ji}(x_j^0) \tag{9}$$

where, $d_{ji}(x_j^0)$ is moving rate for *j*-th membership of *i*-th rule defined by Eq. (5)-(7)

$y^i$ in Eq. (8) is a value calculated by substitution of input $x_1^0, x_2^0, ..., x_j^0, ..., x_m^0$ in the equation of

consequence in Eq. (10).

$$y^i = a_0^i + \sum_{j=1}^{m} a_j^i x_j^0 \qquad (10)$$

The steps to identify the structure and parameters of premise and consequence based on the moving rate are shown in Figure 5.

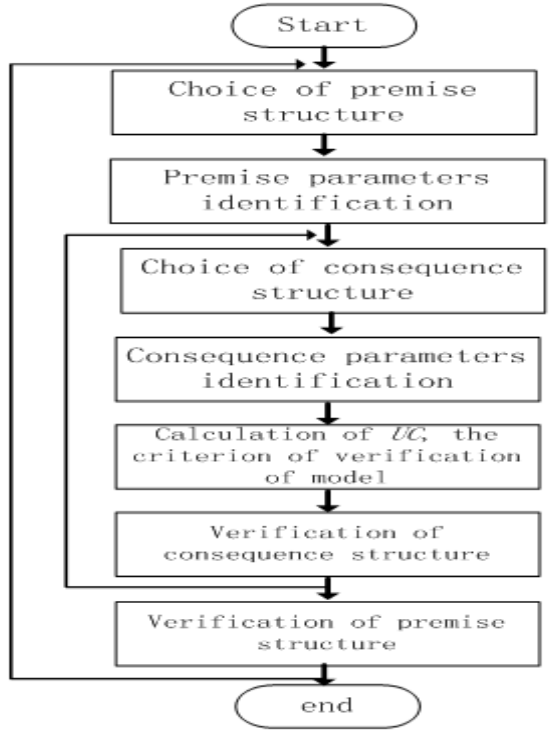

Figure 5. The identification steps of fuzzy model

The proposed method is called T-S modeling or T-S fuzzy identification method based on moving rate.

In this paper, we use unbiased criterion($UC$) which is used as a proper criterion for verification of structure in GMDH [24]: a method for identifying nonlinear models. The basic idea embodied in criterion is as follows: in presence of moderate noise, the parameters of the model with true structure are the least sensitive to the observed data that are used for identifying the parameters. Input and output data for identification can be divided into two sets $N_A$ and $N_B$, and it is feasible to identify the consequence parameters for each set of data separately. Then unbiased criterion $UC$ is calculated as

$$UC = \left( \sum_{i=1}^{n_A} (y_i^{AB} - y_i^{AA})^2 + \sum_{i=1}^{n_B} (y_i^{BA} - y_i^{BB})^2 \right)^{\frac{1}{2}}, \quad (11)$$

Where $n_A$ is number of data set $N_A$, $y_i^{AA}$ is estimated output for data set $N_A$ from the model identified by using data set $N_A$, and $y_i^{AB}$ is a estimated output for data set $N_A$ from the model identified by using data set $N_B$. Therefore, the first clause of Eq. (11) is the difference of estimated output using model $A$ and $B$ for data set $N_A$, and the second is the difference of estimated output for the data set $N_B$.

We must select the structure that $UC$ is minimal. When the structure is proper, the accuracy of the model is the same though we use different data.

### 3.2. Identification method of consequence

Here, an identification method of the structure and parameter of consequence based on the moving rate is proposed.

First, supposing the structure and the parameter of premise are given, and considering the identification method of consequence structure and parameter in Figure 5.

$\hat{d}^i$ is defined as Eq. (12) to simplify Eq. (8).

$$\hat{d}^i = \frac{d^i}{\sum_{i=1}^{n} d^i} \quad (12)$$

Then Eq. (8) by Eq. (9) can be written as follows.

$$\begin{aligned} y^0 = \sum_{i=1}^{n} \hat{d}^i y^i &= a_0^1 \hat{d}^1 + a_1^1 \hat{d}^1 x_1^0 + \cdots + a_m^1 \hat{d}^1 x_m^0 + \\ &+ a_0^2 \hat{d}^2 + a_1^2 \hat{d}^2 x_1^0 + \cdots + a_m^2 \hat{d}^2 x_m^0 + \cdots \\ &+ a_0^n \hat{d}^n + a_1^n \hat{d}^n x_1^0 + \cdots + a_m^n \hat{d}^n x_m^0 \end{aligned} \quad (13)$$

$$\left. \begin{aligned} z_0^i &= \hat{d}^i, & i = 1,2,\cdots,n \\ z_j^i &= \hat{d}^i x_j^0, & j = 1,2,\cdots,m \end{aligned} \right\} \quad (14)$$

$$y^0 = \sum_{i=1}^{n} \hat{d}^i y^i = \sum_{i=1}^{n} (a_0^i z_0^i + a_1^i z_1^i + \cdots + a_m^i z_m^i) \quad (15)$$

In addition, when Eq. (14) is held, Eq. (15) is satisfied. That is, $y^0$ is a linear equation with a variable $z_j^i$ and parameter $a_j^i$. The number of variables or parameters is $n(m+1)$. In case that the structure of consequence is given, identification of parameter $a^i_j$ can be *LSE* (least squares estimator) problem, because Eq. (15) is linear. Of course, the pair of variable $z^i_j$ and output $y^0$ in Eq. (15) is given as data. $z^i_j$ is decided by input $x_j$ and weight $d^i$, and when the premise has been already given, $d^i$ is also given.

On the other hand, to select the consequence variables in structure identification means to obtain several $z^i_j$ among $n(m+1)$ variables, but it does not simply select several $x_j$ among inputs $x_1, ..., x_m$. For example, in case $n=2$, $m=3$, considering that the $z^1_0$, $z^1_2$, $z^2_1$, $z^2_3$ are selected from 8 variables. Then fuzzy model can be described as follows.

$$L^1 : if - then \quad y^1 = a^1_0 + a^1_2 x_2$$
$$L^2 : if - then \quad y^2 = a^2_1 x_1 + a^2_3 x_3$$

As it is shown in Eq. (15), to select the variable $z^i_0$ means that constant clause $a^i_0$ is in linear equation of consequence. That is, the variable $z^i_j$ is introduced to identify the structures of linear model at the same time. If the number of fuzzy rules is $n$, then it is necessary to select at least one $z^i_j$ as the variable for any $i \in [1, n]$. In addition, the selection of consequence variable is done using the variable elimination by *UC*. The method of the variable elimination is the one that firstly making the model by using the whole variables and then eliminating the variables one by one by calculating *UC*. That is, parameter $a^i_j$ must be identified for structure of each consequence whenever eliminating the variables.

Method for the variable elimination is as follows.

First, by applying the whole variables, parameters of consequence are identified, and then the value of *UC* for obtained model is written as $UC_0$. Next, the model with one variable eliminated is made and $UC_1$ is calculated. In this case, the number of variables is $n(m+1)$, so the number of models with one variable eliminated is also $n(m+1)$. If $UC_1$ is greater than $UC_0$, we should insert the variable again and eliminate another one, and should newly calculate $UC_1$. If $UC_1$ is smaller than $UC_0$, then eliminating other variable again, making the model and calculating $UC_2$. If the value of $UC_{k-1}$ is not smaller than $UC_k$, in the case that $k$ variables are eliminated, the structure of the consequence for $UC_k$ becomes an optimal one.

*3.3. Identification of premise*

The main difficulty in fuzzy modeling is identification of premise. The identification step of premise consists of the identification steps for the structure and the parameters. The identification of structure consists of selection of fuzzy variables for premise and fuzzy partition for variable space.

Before considering the identification for premise structure, firstly consider the parameter identification where the premise parameter is the coordinate of *x*-axis $a_1$, $a_2$, $a_3$, $a_4$ characterizing the membership function (Figure 6).

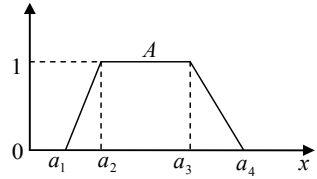

Figure 6. The parameter for the premise membership function

When the parameter values for all premise fuzzy variables are given, the output $y^0$ for weight $d^i$ is obtained. Under this condition, the optimal structure and the consequence parameters are identified.

In order to decide the premise parameters, searching minimal parameter value is needed. To avoid the complexity of computation, there is no need to select the consequence structure. In this case, level 0 is used. That means all the input variables in premise parameter identification. Then, the consequence parameter can be decided using *LSE*. Namely, $a_j^i$ is obtained by using input-output data in Eq. (15).

The standard for premise parameter identification is to select the one that the sum of square of output error in Eq. (15) is minimum or the weight coefficient is maximum. In other words, the objective function for premise parameter identification is to minimize the sum of square of output error in Eq. (15) or to maximize the weight coefficient. These do not really depend on the consequence parameter $a_j^i$. In general, the parameters by calculating the differentiation of objective function may be estimated. However, in Eq. (15), there exists non-differential operation for absolute and min operation. So it is better to use the simplex algorithm [25] that does not need the possibility of differential of objective function in identification of premise parameter.

## 3.4. Identification of consequence parameter

Represent Eq. (8) in the form of matrix.

Refer $p$ the number of input-output pairs, and $Z \in R^{p \times n(m+1)}$ is a matrix where, $Z$ is as follows

$$Z = \begin{pmatrix} z_0^{1\,(1)} & \cdots & z_m^{1\,(1)} & z_0^{2\,(1)} & \cdots & z_m^{2\,(1)} & \cdots & z_0^{n\,(1)} & \cdots & z_m^{n\,(1)} \\ z_0^{1\,(2)} & \cdots & z_m^{1\,(2)} & z_0^{2\,(2)} & \cdots & z_m^{2\,(2)} & \cdots & z_0^{n\,(2)} & \cdots & z_m^{n\,(2)} \\ \cdots & \cdots & \cdots & \cdots & \cdots & \cdots & \cdots & \cdots & \cdots & \cdots \\ z_0^{1\,(p)} & \cdots & z_m^{1\,(p)} & z_0^{2\,(p)} & \cdots & z_m^{2\,(p)} & \cdots & z_0^{n\,(p)} & \cdots & z_m^{n\,(p)} \end{pmatrix} \quad (16)$$

where the element of matrix $Z$, $z_j^{i(k)}$ ($j=0, \ldots, m$, $i=1, \ldots, n$, $k=1, \ldots, p$) is obtained by substituting the $k^{th}$ input data pair in Eq. (14). That is, the model can be expressed in the form of matrix operation like Eq. (17), and the identification problem for fuzzy model results in the problem obtaining the parameter vector $A \in I^{n(m+1) \times 1}$ in linear system expressed as Eq. (17)

$$y = ZA + \varepsilon \quad (17)$$

where, $y = [y_1, y_2, \cdots, y_p]^T$ is a set for $p$ output data, $\varepsilon = [\varepsilon_1, \varepsilon_2, \ldots, \varepsilon_p]^T$ is an error vector memorizing the corresponding errors. This expresses the model error, additional noise or uncertainty.

Parameter vector $A$ in Eq. (17) can be obtained by minimizing the norm of $\varepsilon$. The minimization of $\varepsilon$ norm is *LSE* problem and it is the problem obtaining $\hat{A}$ expressed as Eq. (18)

$$\hat{A} = \min_A \|ZA - y\| \quad (18)$$

where, the symbol $\|\ \|$ means norm.

Applying *LSE* to Eq. (18), $\hat{A}$, estimation value of $A$, is obtained as follows.

$$\hat{A} = (Z^T Z)^{-1} Z^T y \quad (19)$$

Here, if $rank(Z) = \min\{p, n(m+1)\}$ is held for the regression matrix $Z$, then the solution of Eq. (18) is only one and can be obtained using Eq. (19).

If $rank(Z) < \min\{p, n(m+1)\}$, then the determinant of $Z^T Z$ is 0 and $\hat{A}$ can not be obtained by using Eq. (19). However, the type of premise membership to estimate is *s*-type, *z*-type and *trapezoidal* type and it is high probable to have the same membership function for the different input values. So rank shortage problem may be occurred. That is, the matrix $Z$ may have the same rows or columns, and in this case, the determinant of $Z$ became zero. Therefore, we may not identify the parameters using the matrix operation method with Eq. (19).

To solve this problem, we use the singular value decomposition (SVD) algorithm of regressive matrix [26]. We overcome the rank shortage arising in the consequence parameter identification by introducing the SVD algorithm for fuzzy model identification.

**4. Experiment data of nonlinear object and result analysis**

In order to verify the proposed method, following nonlinear system is used [22].

$$y = (1+x_1^{0.5}+x_2^{-1}+x_3^{-1.5})^2 \qquad (20)$$

For this system, *Kondo* identified a model by using the improved GMDH method, which can construct a polynomial models with no integer degree. Table I shows input-output data pair shown in paper [27].

As shown in table I, the number of data pairs is 40. Among them, the number 1-20 are data for identification, and the number 21-40 are for verification of the model. The identified model [22] is

$$y = -3.1+5.2x_1^{0.5118}x_2^{-0.3044}+3.8x_1^{0.4456}x_3^{-0.3371}+10.2x_2^{-0.3174}x_3^{-0.5879} . \qquad (21)$$

Table I input-output data used in experiment

| № | $x_1$ | $x_2$ | $x_3$ | $x_4$ | $y$ | № | $x_1$ | $x_2$ | $x_3$ | $x_4$ | $y$ | № | $x_1$ | $x_2$ | $x_3$ | $x_4$ | $y$ |
|---|---|---|---|---|---|---|---|---|---|---|---|---|---|---|---|---|---|
| 1 | 1 | 3 | 1 | 1 | 11.11 | 15 | 1 | 1 | 3 | 5 | 10.19 | 29 | 5 | 5 | 5 | 1 | 12.43 |
| 2 | 1 | 5 | 2 | 1 | 6.521 | 16 | 5 | 3 | 2 | 5 | 15.39 | 30 | 5 | 1 | 4 | 1 | 19.02 |
| 3 | 1 | 1 | 3 | 5 | 10.19 | 17 | 5 | 5 | 1 | 1 | 19.68 | 31 | 1 | 3 | 3 | 5 | 6.38 |
| 4 | 1 | 3 | 4 | 5 | 6.043 | 18 | 5 | 1 | 2 | 1 | 21.06 | 32 | 1 | 5 | 2 | 5 | 6.521 |
| 5 | 1 | 5 | 5 | 1 | 5.242 | 19 | 5 | 3 | 3 | 5 | 14.15 | 33 | 1 | 1 | 1 | 1 | 16 |
| 6 | 5 | 1 | 4 | 1 | 19.02 | 20 | 5 | 5 | 4 | 5 | 12.68 | 34 | 1 | 3 | 2 | 1 | 17.219 |
| 7 | 5 | 3 | 3 | 5 | 14.15 | 21 | 1 | 1 | 5 | 1 | 9.545 | 35 | 1 | 5 | 3 | 5 | 5.724 |
| 8 | 5 | 5 | 2 | 5 | 14.36 | 22 | 1 | 3 | 4 | 1 | 6.043 | 36 | 5 | 1 | 4 | 5 | 19.02 |
| 9 | 5 | 1 | 1 | 1 | 27.42 | 23 | 1 | 5 | 3 | 5 | 5.724 | 37 | 5 | 3 | 5 | 1 | 13.39 |
| 10 | 5 | 3 | 2 | 1 | 15.39 | 24 | 1 | 1 | 2 | 5 | 11.25 | 38 | 5 | 5 | 4 | 1 | 12.68 |
| 11 | 1 | 5 | 3 | 5 | 5.724 | 25 | 1 | 3 | 1 | 1 | 11.11 | 39 | 5 | 1 | 3 | 5 | 19.61 |
| 12 | 1 | 1 | 4 | 5 | 9.766 | 26 | 5 | 5 | 2 | 1 | 14.35 | 40 | 5 | 3 | 2 | 5 | 15.39 |
| 13 | 1 | 3 | 5 | 1 | 5.870 | 27 | 5 | 1 | 3 | 5 | 19.61 | | | | | | |
| 14 | 1 | 5 | 4 | 1 | 5.406 | 28 | 5 | 3 | 4 | 5 | 13.65 | | | | | | |

The fuzzy model using the identification algorithm in Figure 5 based on data shown in table I is identified. The variable $x_3$ and $x_2$ as fuzzy premise variables are selected, in the case that the number of rules is four, the fuzzy set is divided into five types as shown in Figure 7.

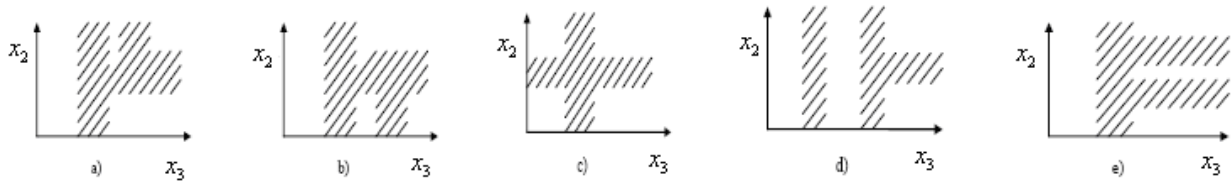

Figure 7. Five fuzzy partitions in the case that the number of rules is four

Based on above experiment results, we can describe as follows.

(Stage 1) In case that the number of rules is one, the result of fuzzy modeling is $y = 11.13+2.02x_1- 1.63x_4$ for data of table I and $UC=3.82$. $UC$ in [28] is 3.8 and a linear model is obtained as follows.

$$y =15.3+1.97x_1-1.35x_2-1.57x_3 \tag{22}$$

(Stage 2) In case that the number of rules is two, the result of fuzzy modeling is divided into four parts according to which variable are used as premise variable. $UC$ in each case is shown in table II.

Table II Comparison of $UC$ in case the number of rules of two

| UC | $x_1$ | $x_2$ | $x_3$ | $x_4$ |
|---|---|---|---|---|
| Existing method [22] | 5.4 | 3.5 | 3.3 | 4.6 |
| Proposed method | 4.11 | 4.04 | 3.72 | 5.78 |

Because we select valuables $x_1$ and $x_4$ as premise variables in stage 2, UC is large. Therefore, these are not selected in later stage as premise variables. UC, minimum in stage 2 of 3.72 is less than 3.82 in stage 1, so it passes to the next stage.

(Stage 3) when the fuzzy set is divided at $x_3$-$x_2$ spaces, we can consider three fuzzy partitions such as in Figure 8. The value of UC in this case is shown in table III. The minimum UC is c) of Figure 8.

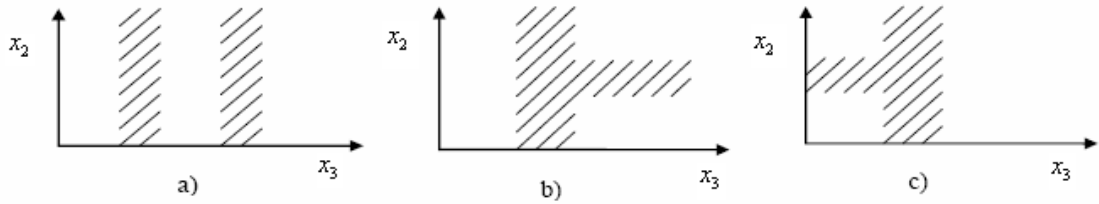

Figure 8. Fuzzy partition in the case that the number of rule is three

Table III comparison of UC in case of the number of rule of three

| UC | a) | b) | c) |
|---|---|---|---|
| Existing method [22] | 4.2 | 2.8 | 3.3 |
| Proposed method | 3.49 | 0.92 | 3.51 |

Since the value of b) in table III is minimal, this structure is selected. The minimum UC is 0.92 and it is less than 3.72 in stage 2, so it passes to the stage 4.

(Stage 4) When we divide fuzzy set in $x_3$-$x_2$ spaces reserving the structure c) in Figure 8, we can consider five fuzzy partitions as in Figure 7. UC in this case are shown in table IV. Minimum UC is 1.24 and is greater than 0.92 in stage 3, so we select the modeling result in stage 3 as result.

Table IV comparison of UC in case of the number of rules of four

| UC | a) | b) | c) | d) | e) |
|---|---|---|---|---|---|
| Existing method [22] | 5.7 | 6.2 | 3.4 | 6.7 | 7.2 |
| Proposed method | 1.24 | 1.74 | 1.74 | 1.49 | 1.67 |

Comparing the accuracy of identified fuzzy model with various models.

In paper [22], Eq. (23) is used as index of evaluating the model accuracy $E$.

$$E = \frac{1}{N}\sum_{i=1}^{N}\frac{|y_i - \overline{y}_i|}{y_i} \times 100[\%] \tag{23}$$

Table V shows the comparison with various models.

Table V the error and $UC$ of each model

| Index / model | $E_{1(\%)}$ | $E_{2(\%)}$ | $UC$ |
|---|---|---|---|
| Linear model (Eq.(22) ) | 12.7 | 11.1 | 3.8 |
| Improved GMDH model(Eq.(21) ) | 4.7 | 5.7 | × |
| Sugeno's fuzzy model 1[22] | 1.5 | 2.1 | 2.1 |
| Sugeno's fuzzy model 2[22] | 0.59 | 3.4 | 3.4 |
| Proposed fuzzy model | 0.258 | 0.267 | 1.24 |

In table V, fuzzy model *1* is identified in step 3, and fuzzy model *2* is in the third of stage 4. In [22], evaluation standard $UC$ is used to evaluate the accuracy of model. In table V, $E_1$ is obtained using identified model for identification data, and $E_2$ is the result estimated the output using identified model for evaluation data. In table V, we can see that in *Sugeno*'s fuzzy model *1* the accuracy is much higher than that of the model by GMDH. The difference between $E_1$ and $E_2$ is not large for the same model. In other words, fuzzy model *2* is better than fuzzy model *1* for the value of $E_1$. However, for the evaluation data, $E_2$=3.4, and the estimation accuracy is worse than $E_2$=2.1 of fuzzy model *1*. And $UC$ is also greater than model 2, therefore we can see that the fuzzy model *1* is better than fuzzy model *2*.

The results of comparing the proposed method with existing methods [22] are as follows.

$E_1$ is 0.258, $E_2$ is 0.267 and $UC$ is 1.24 in table V in the proposed method. Therefore, we may concluded that the accuracy of the proposed method is higher than fuzzy models *1* and *2*.

## 5. Application to prediction model of precipitation and security situation

### 5.1. Experiment data for precipitation

In this paper, mean square error (MSE) is used as the performance index in fuzzy modeling, which is defined as following

$$MSE = \frac{1}{k}\sum_{i=1}^{k}(y_k - \hat{y})^2. \tag{24}$$

where $y_k$ is $k$-th original system output, $\hat{y}$ is $k$-th data of model output.

The measured data from 1952 to 1977 by China Tianjin city Weather Service is as table VI[29].

Table VI. Measured data for precipitation forecast model[29]

| YEAR | FACTOR 1 | FACTOR 2 | PRECIPITATION (MM) | YEAR | FACTOR 1 | FACTOR 2 | PRECIPITATION (MM) |
|---|---|---|---|---|---|---|---|
| 1952 | 0.73 | -5.28 | 283 | 1965 | 0.46 | -14.68 | 348 |
| 1953 | -2.08 | 5.18 | 647 | 1966 | -2.31 | -1.36 | 644 |
| 1954 | -3.53 | 10.23 | 731 | 1967 | 0.2 | -5.43 | 431 |
| 1955 | -3.31 | 4.21 | 561 | 1968 | 3.46 | -19.85 | 179 |
| 1956 | 0.53 | -2.46 | 467 | 1969 | 0.08 | 8.59 | 615 |
| 1957 | 2.33 | 7.32 | 399 | 1970 | 1.46 | 7.26 | 433 |
| 1958 | -0.32 | -10.81 | 315 | 1971 | 0.24 | -1.1 | 401 |
| 1959 | -2.35 | 3.85 | 521 | 1972 | 0.89 | -16.94 | 206 |
| 1960 | -0.95 | 2.74 | 472 | 1973 | -0.5 | 10.46 | 639 |
| 1961 | -0.64 | 6.0 | 536 | 1974 | 2.15 | -10.06 | 418 |
| 1962 | 0.92 | 0.65 | 385 | 1975 | -0.89 | 12.11 | 570 |
| 1963 | 2.98 | -11.83 | 259 | 1976 | 1.4 | -6.26 | 415 |
| 1964 | -0.85 | -2.3 | 657 | 1977 | -0.59 | 7.15 | 796 |

The experiment result in proposed method is shown in Figure 9, 10.

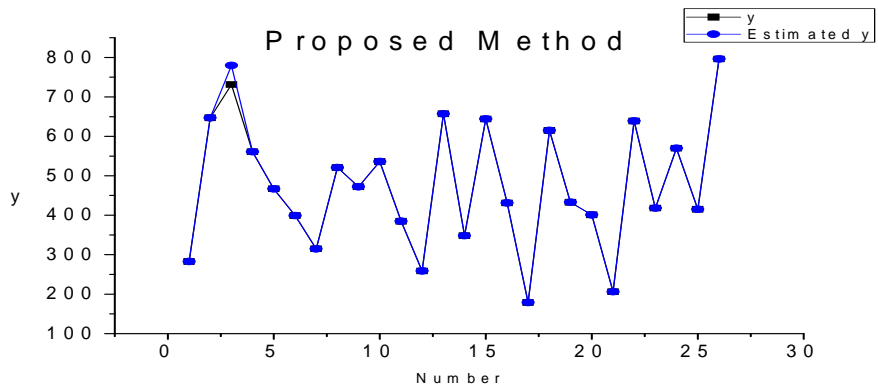

Figure 9. Experiment result in proposed method

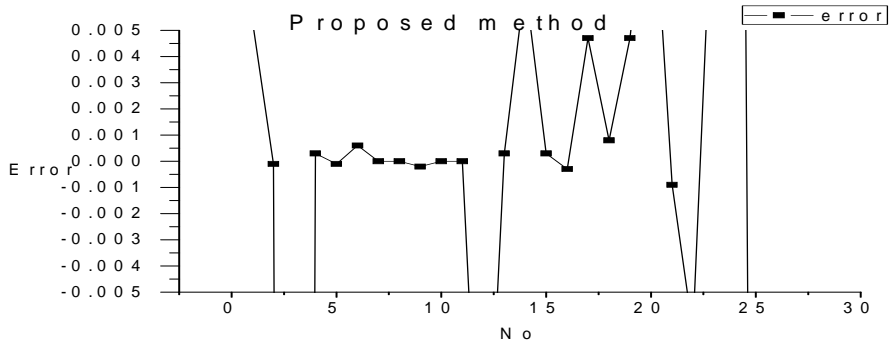

Figure 10. Errors in proposed method (MSE = 0.00094864)

The fuzzy model for this structure is as follows.

*if $u_1(t-4)$ is [-2.808 5 -0.608 28]s then*

$y$=-2 387.320 5+2.832 8 $y(t-1)$+306.074 8 $u_1(t-3)$-711.693 2 $u_1(t-4)$+126.809 8 $u_2(t-3)$-45.081 2 $u_2(t-4)$

*if $u_1(t-4)$ is [-1.059 9 -0.321 17]B, $u_1(t-3)$ is [-2.558 9 0.617 38]S then*

$y$= -1 015.134 3+2.508 $y(t-1)$-14.411 5 $u_1(t-3)$+445.250 7 $u_1(t-4)$-109.179 1 $u_2(t-3)$-54.115 1 $u_2(t-4)$

*if $u_1(t-4)$ is [-1.059 9 -0.321 17 0.015 475 1.367 3]m, $u_1(t-3)$ is [-1.963 6 1.864 8]b then*

$y$ =2 597.007 8-2.369 8 $y(t-1)$-268.836 $u_1(t-3)$ -2 437.606 8 $u_1(t-4)$+20.311 4 $u_2(t-3)$-119.882 1 $u_2(t-4)$

*if $u_1(t-4)$ is [0.439 2 1.939 4]S, $u_1(t-3)$ is [-1.963 6 1.864 8]B then*

$y$=-872.636 5+2.055 2 $y(t-1)$+748.048 9$u_1(t-3)$-845.325 6$u_1(t-4)$-135.626$u_2(t-3)$-84.747$u_2(t-4)$

In this model, $u1(t-i)(i=1,2,3...)$ is input at *t-i* of the first input variable, and B is s-type membership(means large), S is z-type membership(means small) and M is trapezoidal membership function(means Medium). And the values in [...] are the axis values where the grade of membership function is either 0 or 1.

### *5.2. Experiment data for security situation*

The proposed method is described through the experiment for security situation and its analysis (table VII).

Table VII. Observation data for security situation

| Factor 1 | Factor 2 | Factor 3 | Security state value | Factor 1 | Factor 2 | Factor 3 | Security state value |
|---|---|---|---|---|---|---|---|
| 1 | 2 | 7 | 13.792 | 2 | 1 | 8 | 16.354 |
| 1 | 6 | 9 | 14.783 | 9 | 4 | 2 | 94.707 |
| 5 | 1 | 9 | 37.333 | 8 | 4 | 8 | 77.354 |
| 5 | 2 | 9 | 37.748 | 6 | 2 | 3 | 48.992 |
| 7 | 6 | 8 | 62.803 | 8 | 3 | 7 | 77.11 |
| 4 | 2 | 1 | 29.414 | 6 | 4 | 5 | 49.447 |
| 8 | 6 | 6 | 77.858 | 9 | 4 | 6 | 94.408 |
| 6 | 9 | 3 | 50.577 | 4 | 1 | 6 | 28.408 |
| 4 | 2 | 6 | 28.822 | 4 | 8 | 4 | 30.328 |
| 5 | 2 | 1 | 38.414 | 2 | 6 | 7 | 17.827 |
| 1 | 1 | 4 | 13.5 | 2 | 4 | 9 | 17.333 |
| 3 | 4 | 7 | 22.378 | 1 | 3 | 7 | 14.11 |
| 1 | 5 | 2 | 14.943 | 5 | 6 | 7 | 38.827 |
| 1 | 8 | 9 | 15.162 | 2 | 7 | 8 | 17.999 |
| 9 | 1 | 1 | 94 | 7 | 8 | 6 | 63.237 |
| 9 | 3 | 7 | 94.11 | 4 | 9 | 6 | 30.408 |
| 3 | 9 | 9 | 23.333 | 1 | 4 | 6 | 14.408 |
| 4 | 7 | 5 | 30.093 | 3 | 9 | 1 | 24 |
| 3 | 4 | 9 | 22.333 | 3 | 7 | 6 | 23.054 |
| 5 | 7 | 1 | 39.646 | 1 | 7 | 3 | 15.223 |
| 4 | 7 | 7 | 30.024 | 1 | 1 | 1 | 14 |
| 7 | 8 | 9 | 63.162 | 5 | 7 | 7 | 39.024 |
| 2 | 4 | 4 | 17.5 | 8 | 1 | 7 | 76.378 |
| 1 | 8 | 4 | 15.328 | 4 | 6 | 8 | 29.803 |
| 6 | 7 | 1 | 50.646 | 2 | 4 | 7 | 17.378 |
| 4 | 1 | 1 | 29 | 7 | 4 | 8 | 62.354 |
| 9 | 4 | 1 | 95 | 1 | 4 | 1 | 15 |
| 6 | 2 | 1 | 49.414 | 6 | 6 | 5 | 49.897 |
| 3 | 1 | 8 | 21.354 | 4 | 1 | 9 | 28.333 |
| 5 | 4 | 5 | 38.447 | 3 | 2 | 4 | 21.914 |

Experiment result in proposed method is shown in Figure 11, 12.

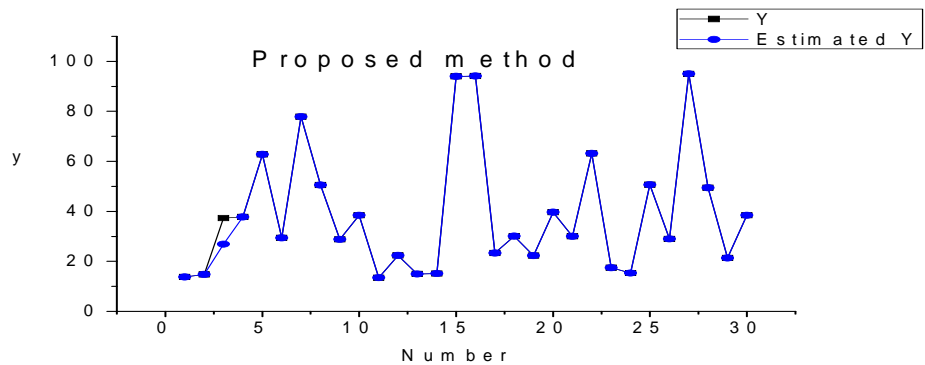

Figure 11. Experiment result for security situation by proposed method

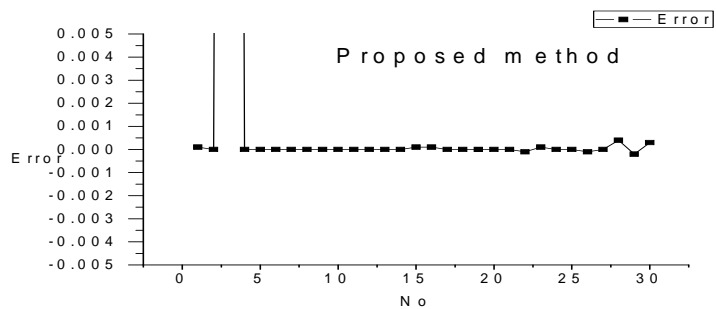

Figure 12. Errors in proposed method (MSE = 0.000 333 345)

The fuzzy model for this structure is as follows.

*if* $u_2(t-3)$ *is* [-0.289 99  7.2612]$s$ $u_2(t-4)$ *is* [-1.280 5  6.316]$s$ *then*

$y$=13.042 7+0.937 17 $y(t-1)$-10.073 8 $u_1(t-4)$+5.360 1$u_1(t-4)$+39.522 3$u_2(t-3)$-7.043 4$u_2(t-4)$-11.321 8$u_3(t-3)$-8.76$u_3(t-4)$

*if* $u_2(t-3)$ *is* [-0.289 99  7.261 2]$s$ $u_2(t-4)$ is [1.909 4   2.157 7]$b$ *then*

$y$=15.283-1.0871$y(t-1)$ +2.275 4 $u_1(t-3)$+9.543 1 $u_1(t-4)$-12.837 8 $u_2(t-3)$+2.637 3 $u_2(t-4)$+7.075 5 $u_3(t-3)$-0.575 95 $u_3(t-4)$

*if* $u_2(t-3)$ *is* [3.768 18  10.553 3]$b$ $u_2(t-4)$ *is* [-1.280 5  6.316]$s$ *then*

$y$=13.567 9-3.900 1 $y(t-1)$+ 23.812 9 $u_1(t-3)$-39.493 1 $u_1(t-4)$-20.561 2 $u_2(t-3)$+5.133 5 $u_2(t-4)$+11.279 8 $u_3(t-3)$+61.300 3 $u_3(t-4)$

*if* $u_2(t-3)$ *is* [3.7681 8  10.553 3]$b$ $u_2(t-4)$ is [1.909 4 2.157 7]$b$ *then*

$y$=21.158+2.187 $y(t-1)$- 7.349 6 $u_1(t-3)$-19.195 4 $u_1(t-4)$+24.186 8 $u_2(t-3)$-4.018 $u_2(t-4)$-4.907 9 $u_3(t-3)$+0.428 25 $u_3(t-4)$

Table VIII shows the performance comparison with existing models.

Table VIII. Comparison with existing models

| Model | MSE |
| --- | --- |
| Sugeno and Yasukawa [30] | 0.079 |
| Nozaki et al. [31] | 0.008 5 |
| Lee et al. [32] | 0.014 8 |
| Kung and Su [33] | 0.019 6 |
| Li et al. [34] | 0.008 5 |
| Our model (precipitation forecast) | 0.000 948 64 |
| Our model (security situation) | 0.000 333 345 |

The comparison between existing methods and the proposed method is as follows. As shown in table VIII, the MSE of proposed method is 0.000 948 64 for precipitation forecast model and the MSE is 0.000 333 345 for security situation model. Therefore, proposed method improves the accuracy of model compared with existing methods and the convergence to the optimal solution is better than existing methods. The reason is because the proposed method is a reasoning method that different reasoning results from different membership function and input information are obtained [35, 36, 37].

## 6. Conclusion

In this paper, the limitation of identification method based on existing matching degree is considered, the moving rate of *s*-type, *z*-type and *trapezoidal* membership function are defined and the fuzzy modeling method based on it is proposed. Then, the identification method using moving rate is proposed, applying simplex algorithm and SVD method to solve non-differential operation and the rank shortage problem. Then, the proposed method is applie d to nonlinear system and fuzzy modeling of precipitation forecast and security situation prediction. It proves that the model accuracy of proposed method is better thans existing methods.

Acknowledgement

This paper is supported by the National Natural Science Foundation of China (No. 6097015), the Doctor Start Foundation of Liaoning Province (No.2081019). This paper is based on "Structure identification of fuzzy model" of *M.Sugeno* and *G.Kang*'s paper (1988), and improves it.